# Robotic positioning device for three-dimensional printing

*Three-dimensional printing device independent from the size of printed objects*

*Saša Jokić[1], Petr Novikov[2], Stuart Maggs[3], Dori Sadan[4], Shihui Jin[5], Cristina Nan[6]*
*Institute for Advanced Architecture of Catalonia[9], Open Thesis Fabrication, Barcelona, Spain*
*http://www.iaac.net, http://robots.iaac.net*
[1]sasa@sasajokic.com, [2]petr@petrnovikov.com, [3]stuart.maggs@iaac.net, [4]dori.sadan@iaac.net, [5]jin.shihui@iaac.net, [6]cristina.nan@iaac.net, [9]info@iaac.net

**Abstract.** Additive manufacturing brings a variety of new possibilities to the construction industry, extending the capabilities of existing fabrication methods whilst also creating new possibilities. Currently three-dimensional printing is used to produce small-scale objects; large-scale three-dimensional printing is difficult due to the size of positioning devices and machine elements. Presently fixed Computer Numerically Controlled (CNC) routers and robotic arms are used to position print-heads. Fixed machines have work envelope limitations and can't produce objects outside of these limits. Large-scale three-dimensional printing requires large machines that are costly to build and hard to transport. This paper presents a compact print-head positioning device for Fused Deposition Modeling (FDM) a method of three-dimensional printing independent from the size of the object it prints.

## I. Introduction

Additive manufacturing brings a variety of new possibilities within manufacturing, current fabrication technologies are highly standardized. Deviation from standard sizing or elements results in a significant production costs increases. Additive manufacturing can offers the possibility to significantly reduce these costs allowing for greater customized fabrication.

Three-dimensional printing is used predominantly to produce small scale objects; the scaling up of this process creates a number of problems. One problem is that large-scale three-dimensional printing faces is size and configuration of positioning devices and of their machine elements.

This paper focuses only on the Fused Deposition Modeling (FDM) method of three-dimensional printing. Currently fixed Computer Numerically Controlled (CNC) routers and robotic arms are predominantly used in additive manufacturing as print head positioning devices. Objects are printed inside the work envelope of the machine, which leads to the machine being larger than the object it can produce. The use of scaled up versions of CNC routers or robotic arms proves to be economically nonviable due to high costs of machine transportation and size.

It is apparent that the need exists for an additive manufacturing positioning device capable of positioning the print-head during the fabrication process, independent of the objects scale. The described research was directed towards providing such a technique.

The proposed device attaches onto the structure that it has printed, moving along the structure while printing. In such manner the device can cover area far greater than its own size making it independent from the size of the desired printed object.

## III. Background

There are several prior-art devices and methods for additive manufacturing that are used to form three-dimensional objects under the control of a computer program. Four of the most well-known methods are FDM (Fused Deposition Modeling), SLS (Selective Laser Sintering), SLA (Stereolithography), Powerbed and inkjet head 3D printing. These methods of forming three-dimensional objects share one important characteristic: they all produce three-dimensional objects from 3D computer data by creating a set of thin two-dimensional cross-sectional slices of the object, forming the object by laying the cross-sections in an additive set of adhered laminae. Various positioning devices are used in these methods but the deposition strategy is the same.

Current research focuses on positioning devices for FDM method. FDM methods. 'Three-dimensional objects may be produced by depositing repeated layers of solidifying material until the shape is formed. Any material, such as self-hardening waxes, thermoplastic resins, molten metals, two-part epoxies, foaming plastics, and glass, which adheres to the previous layer with an adequate bond upon solidification, may be utilized. Each layer base is defined by the previous layer, and each layer thickness is defined and closely controlled by the height at which the tip of the print-head is positioned in relation to the previous layer'[1].

To overcome the issues presented in relation to the scaling of positioning devices, this research aims to create a compact positioning device independent of the size of the object it prints.

# III. Device description
## Supply and communication

The system consists of the positioning device itself (Fig. 1, 1) thats movement is determined by a controller (Fig. 1, 2); the positioning device is connected via tubes to an external material supply (Fig. 1, 3) and cables to the power supply (Fig. 1, 4). Material cartridges can be stored both externally and on the device itself; power connection comes from an external power supply or batteries mounted within the device. Connection to the controller can be established physically through cables or wireless connection.

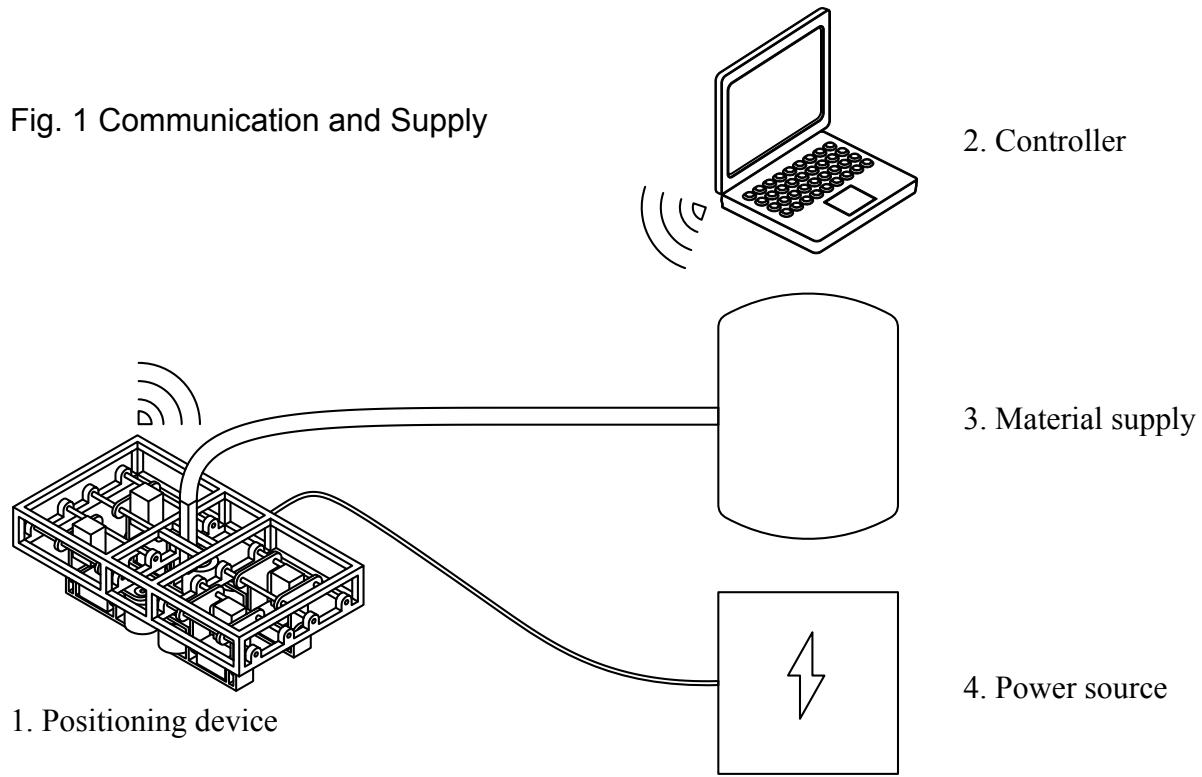

Fig. 1 Communication and Supply

2. Controller

3. Material supply

4. Power source

1. Positioning device

The path of the device, and the parameters of the moving parts during the printing process are inputted to the controller through custom software. The software interprets an object design file generating device control signals based on the computer modeled design. This generation process includes creating a pre-determined series of curves that the device will travel along creating the object.

## Device configuration

The device has a static frame (Fig. 2, 1) incorporating a print-head positioning system (Fig. 2, 11, 12) and device positioning system (legs and feet) (Fig. 2, 4 and 5).

The print-head positioning system comprises of a front-back linear motion system (Fig.2, 11), side linear motion system (Fig. 2, 12), front-back motion actuation system (Fig. 2, 13), side motion actuation system (Fig. 2, 14) and actuators for print-head or multiple print-heads. The connection of print-heads to the material supply is illustrated in (Fig. 2, 15).

The device positioning system consists of feet (Fig. 2, 5) each attached to a further four or more legs (Fig. 2, 4) that are attached to the frame. Legs are connected to the frame via a linear motion system (Fig. 2, 9). Linear motion bearings (Fig. 2, 10) allow smooth movement relative to the frame. This motion can be actuated either by springs or by linear actuators (Fig. 2, 8). Feet are attached to legs via rotary joints that allow rotation relative to the legs. This rotation can be controlled and is actuated by the foot actuation system (Fig. 2, 7).

Wheels (Fig. 2, 3) are mounted on the feet and are moved using the wheel actuation system (Fig. 2, 6). The surface is coated with a durable, flexible material to reduce vibration of the device during movement and to increase its grip to the structure on which the device is attached.

Fig. 2 (a) Device top projection

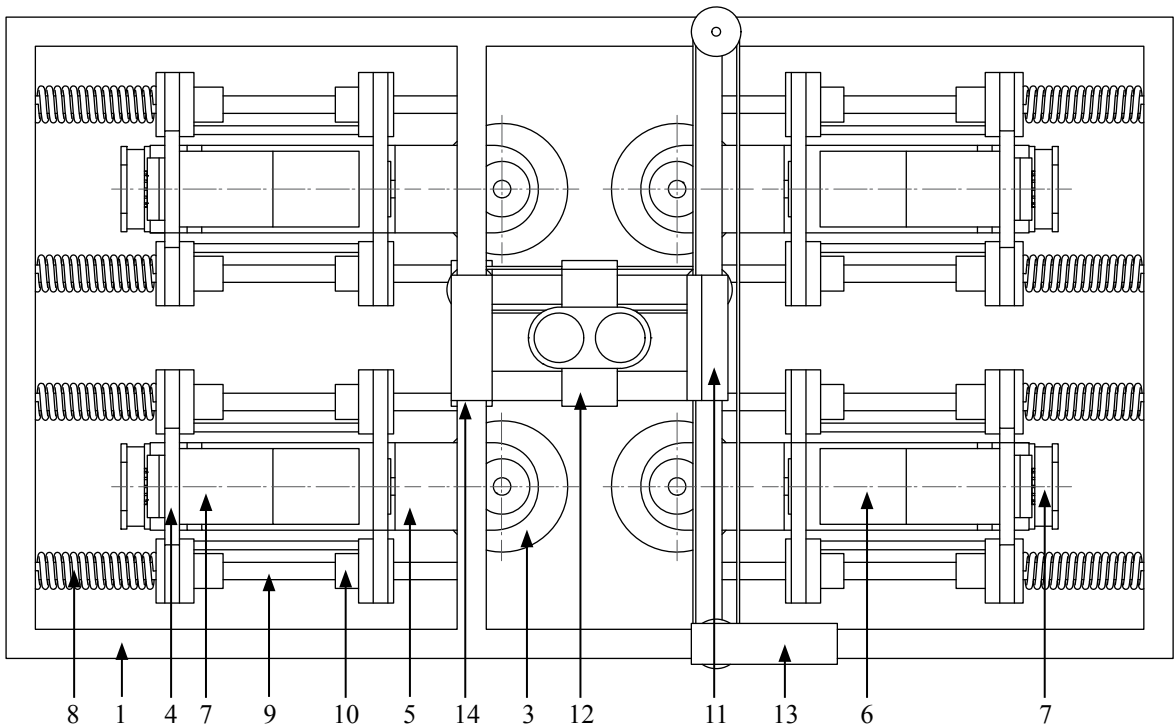

8   1   4  7   9   10   5   14   3   12   11   13   6   7

Fig. 2 (b) Device front projection

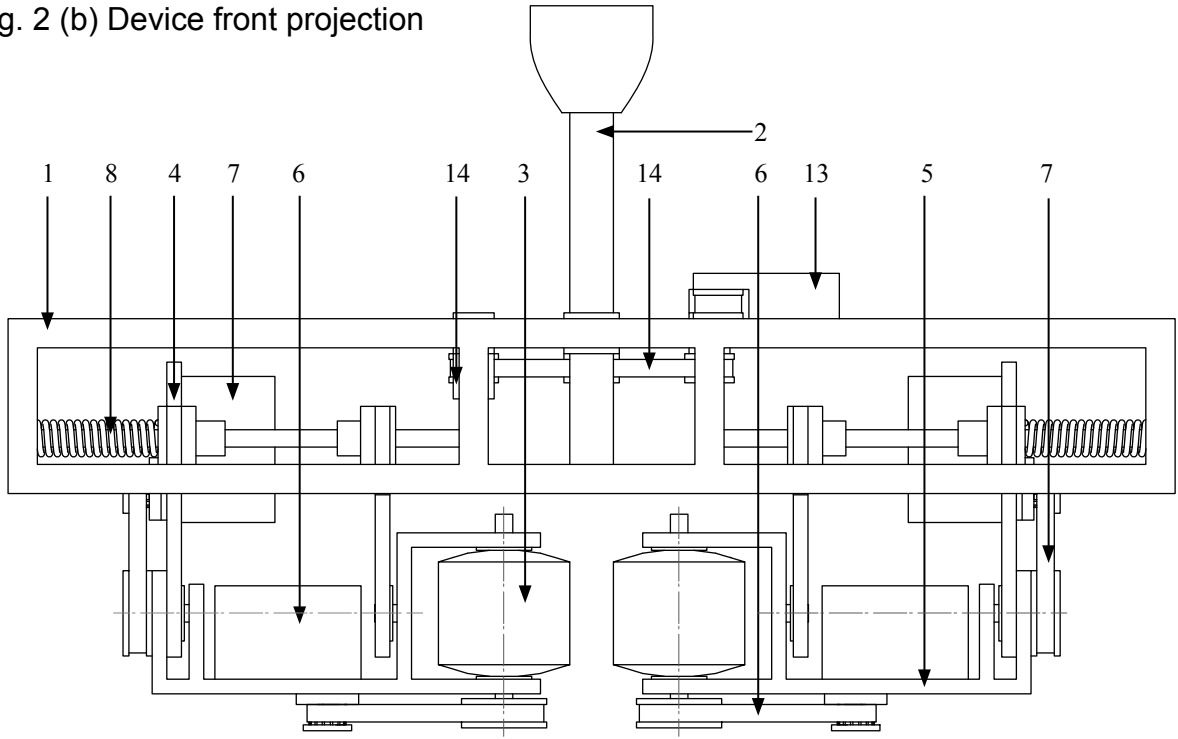

Fig. 2 (c) Device isometric view

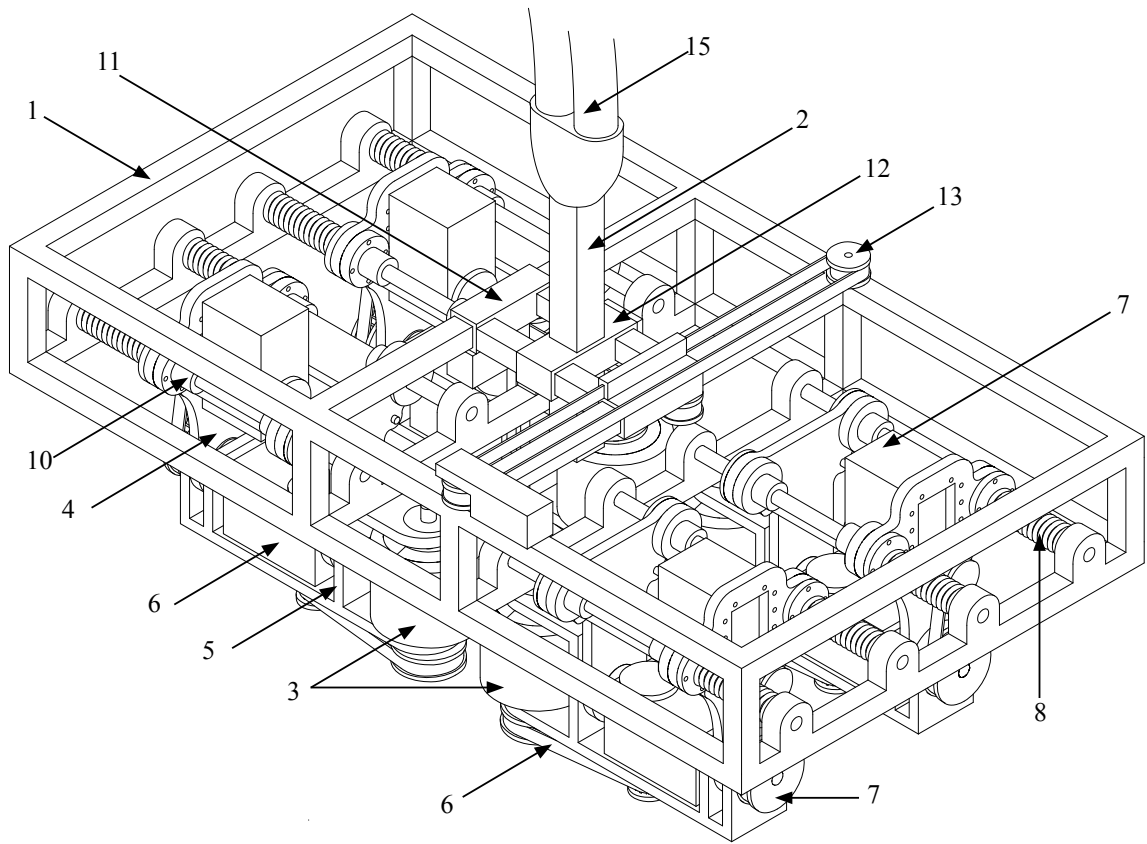

## Device movements

Actuation systems may have various types of actuators and transmissions. In the case actuators require electricity, air pressure, hydraulic or another energy supply they can be connected to such a supply; actuation is then regulated by the controller.

Figure 3 (a) shows the top projection of the device describing wheel motions. The wheels rotate continuously in both directions at varying speeds, depending on the curing time of the material used for printing. Each wheels speed can also can be independently varied, this is necessary when printing curved paths illustrated in (Fig. 4, (a)). In this case the length of the surface on one side of the structure is larger than on the other, resulting in different distances covered by wheels on facing sides.

Figure 3 (b) shows side projections of the device and describes foot rotation relative to the legs. Movement of the feet should be between 0 and 180° degrees of rotation. Their normal position for the devices horizontal movement is such that the axis of wheel rotation is perpendicular to the top surface of the printed layer. To move the device perpendicular to the layer, the feet need to rotate 90°. Varying the angles of rotation can be used to control the position of the device relative to the printed structure. The rotation of the feet can also be used as a steering mechanism during the printing process in order to maintain the correct distance between print-head and the previously printed structure.

Figure 3 (c) shows a front projection of the device and describes the linear motion of legs relative to the frame. This motion is required for the wheels to grip the structure that the device is attached to; facing wheels are pushed towards each other so the distance between them is equal to the thickness of the structure at the point where the wheels are located at a given moment. The surfaces of the wheels connect to the surface of the structure allowing movement relative to the structure. This motion can achieved by uncontrolled actuators such as springs or by controlled linear actuators. Controlled linear actuators increase the ability to manipulate the devices position but it requires more complex software, additional processing and sensors to position wheels correctly.

Figures 3 (d) and (e) show top projections of the device and describe the linear motion of the print-head relative to the frame. Movement of the print head can be along two axis forward-backwards, left and right. The print-head's vertical position can be static relative to the frame, but the distance from the print-head to the structure is defined by the position of the device. The desired distance between the aperture of the print-head and top surface of the structure is the vertical thickness of one printed layer.

Fig. 3 Moving parts

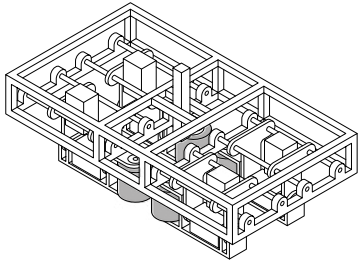
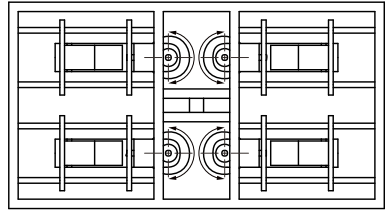

(a) Wheels

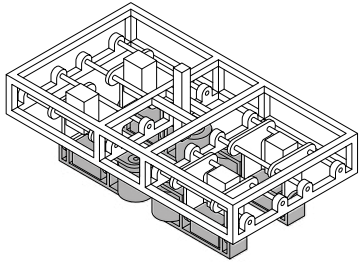
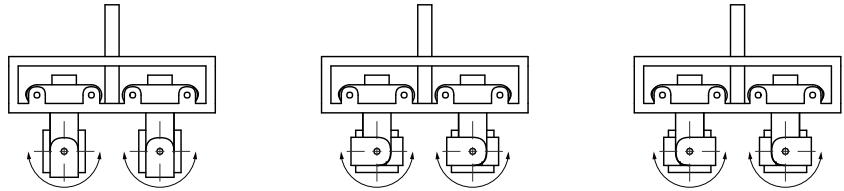

(b) Feet

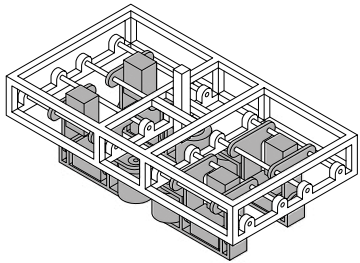
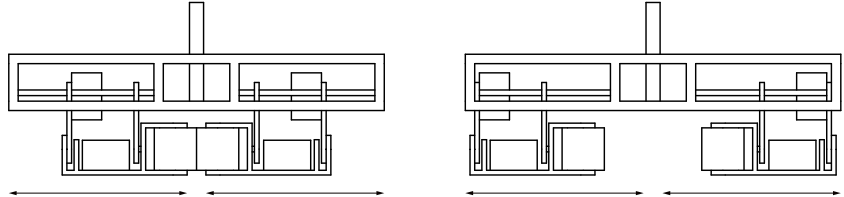

(c) Legs

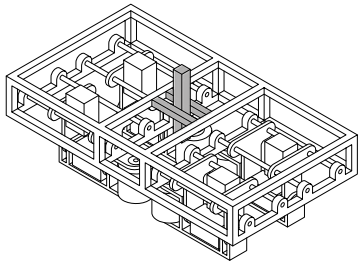
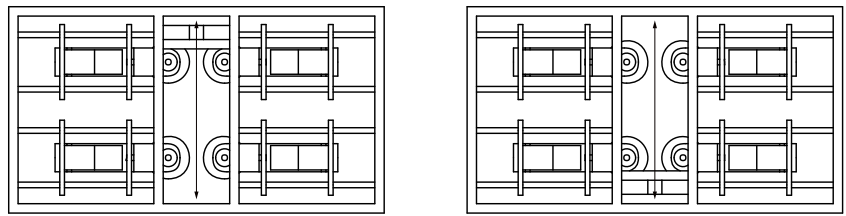

(d) Print-head front-back motion

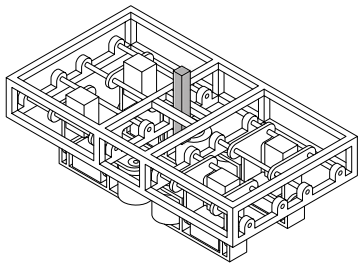
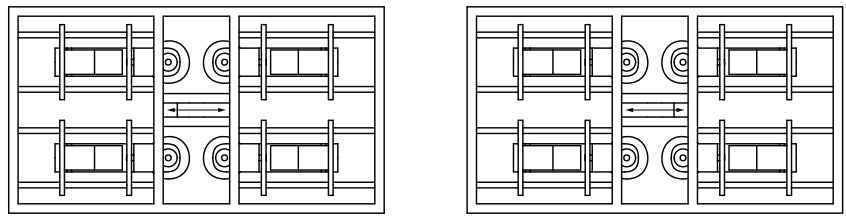

(e) Print-head side motion

# III. Process description

Printing requires a prefabricated footprint structure for the robot to be initially attached to. Such a footprint should be at least the height of the device foot, but ideally more in order for the device not to collide with the ground during printing process. The width of the footprint is ideally as similar to the thickness of printed layers as possible and the shape defined according to the desired final shape of the printed object. It can be both a closed or an open path and is not required to be horizontal, it can also be orientated to surfaces of any angle. The upper surface of the footprint is not required to be flat, but may have various heights (Figure 4 (b)).

Other controllable factors include the ability to increase or decrease the material extrusion speed and the speed of device movement. This relationship defines the dimensions of printed layers and can be changed within a predefined range. This range depends on the range of linear motion system of legs and on the material properties of material being used. For example with a constant extrusion speed of the printing material, the change in speed of the device will decrease or increase the width of the layer. Slowing the device will deposit more material and increase the thickness whilst increasing the speed will reduce the layer width. If the device movement speed is constant, increased material extrusion speed will result in increased layer thickness and decreased material extrusion speed will result in decreased layer width.

## Continuous path

In the case the printing path is closed without openings, the movement of the robot is continuous and front-back print-head motion may not be used. The device is placed on the footprint at a desired position, once material is extruded through the aperture of the print-head and the device starts moving in the desired direction as shown in Figure 4, 1. When the device makes a complete rotation and returns to the starting position having printed the first layer, the feet are rotated perpendicular to the layer moving the device upwards. The distance between print-head aperture and the top surface of previously printed layer is equal to the height of a single layer. This operation is conducted without material extrusion; when the device is ready to print the next layer, the printing process is repeated. This processes is repeated until the structure is complete.

Another way to print without pausing is to use one continuous spiral path instead of multiple layer paths. In this case the wheels are not in their centralized position during the printing process, the device is constantly moving upwards; every complete rotation the device moves up by the height of one layer. This technique may result in a better layer adhesion and better suits the use of liquid printing materials.

Fig. 4 Device movement during continuous printing

(a) Isometric view of device attached to a circular footprint

(b) Isometric view of device attached to a footprint with irregular surface

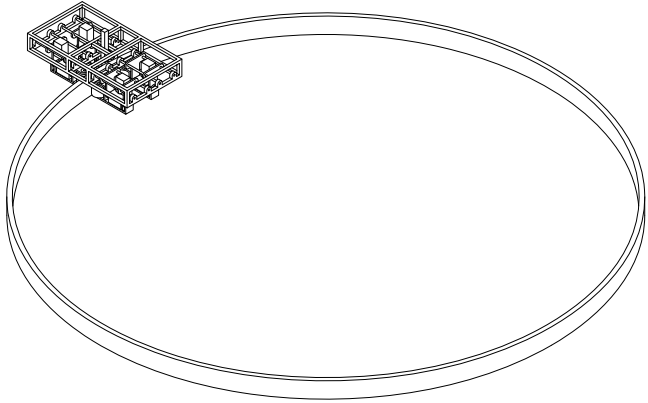
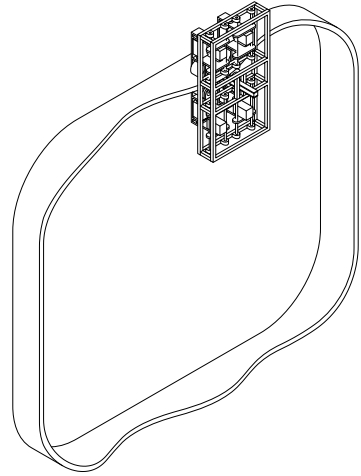

(c) Side view of device during printing process

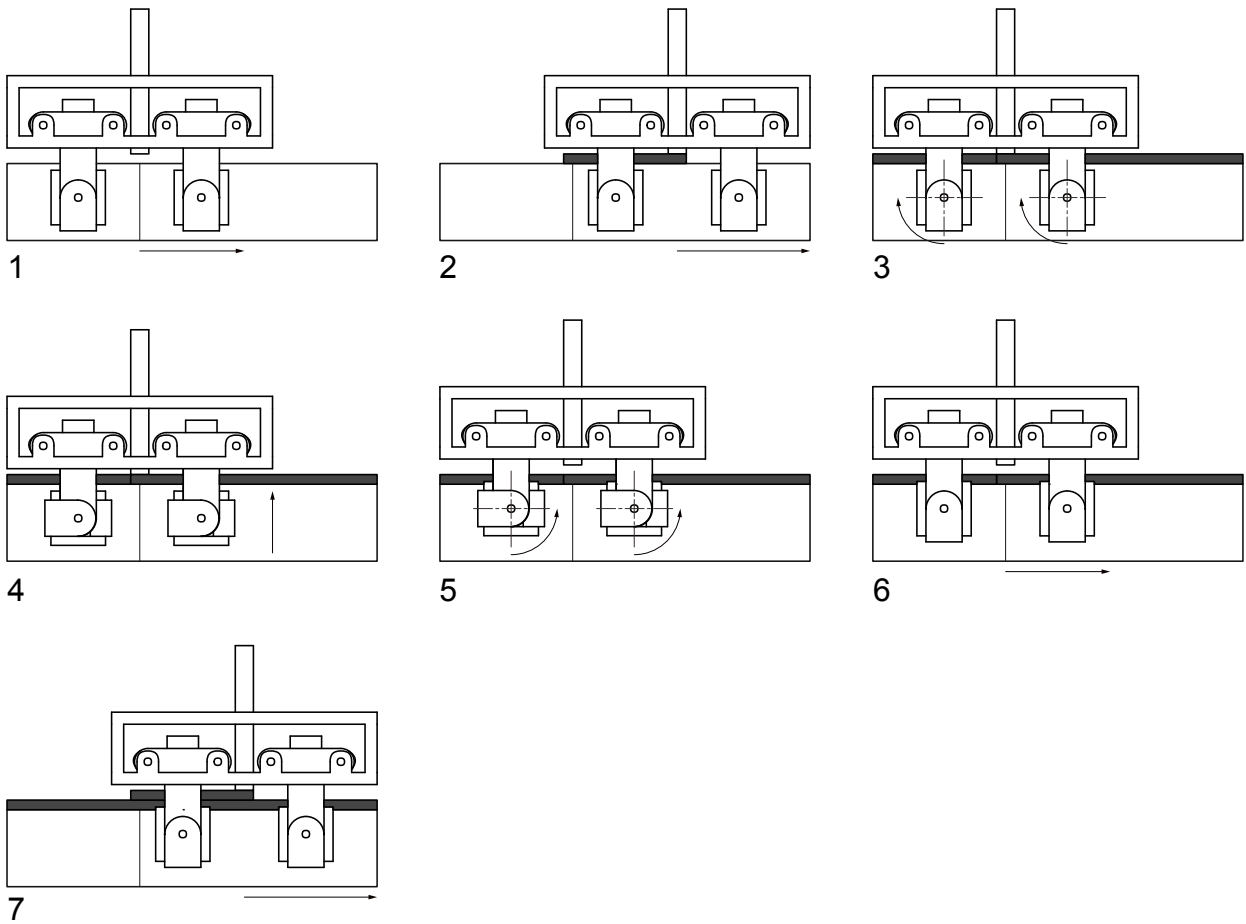

## Discontinuous path

If the printed structure is discontinuous, above mentioned processes are not applicable; once the device reaches the end of the structure it has to move in the reverse direction. Figure 5 shows possible printed discontinuous structures, Figure 6 outlines the process for the printing of discontinuous structures.

The device attaches to the footprint at the starting point the desired path; the distance from the top surface of the footprint to the aperture of the print-head is equal to the height of one layer; the print-head is moved above the starting point at one end of the desired path. Once material is extruded through the aperture of the print-head, the print-head moves towards the center of the device. Once it has reached the center, the device starts moving towards the other end of the path. Upon the device reaching the end of the path, the print-head moves towards the end point completing the first layer. To create the distance of one layer between the print-head aperture and the previously printed layer, the feet are rotated perpendicular to the layer moving the device upwards. This operation is completed without material extrusion, when the device is ready to print the next layer, material is extruded through the print-head. The print-head starts moving towards the center of the device, upon reaching the center, the device starts moving towards the opposite side. When the device reaches the end of the path the print-head moves towards the starting point finishing the second layer and the device lifts itself to the height of one layer. This process is repeated until the structure is completed.

Fig. 5 Isometric view of different discontinuous structures

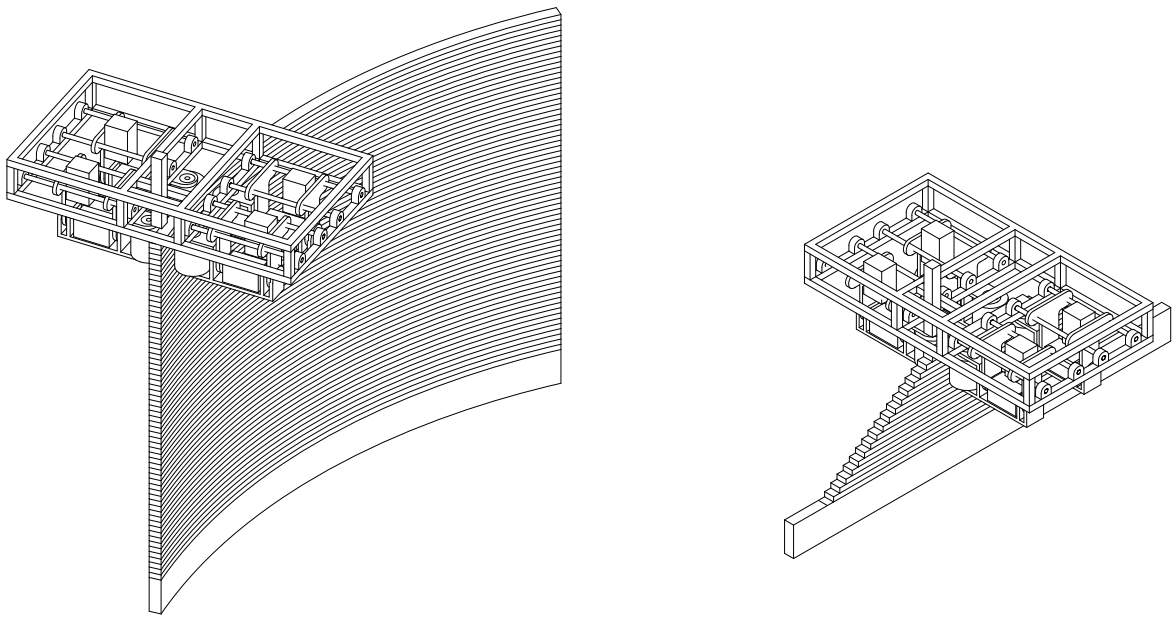

Fig. 6 Device movement during discontinuous printing

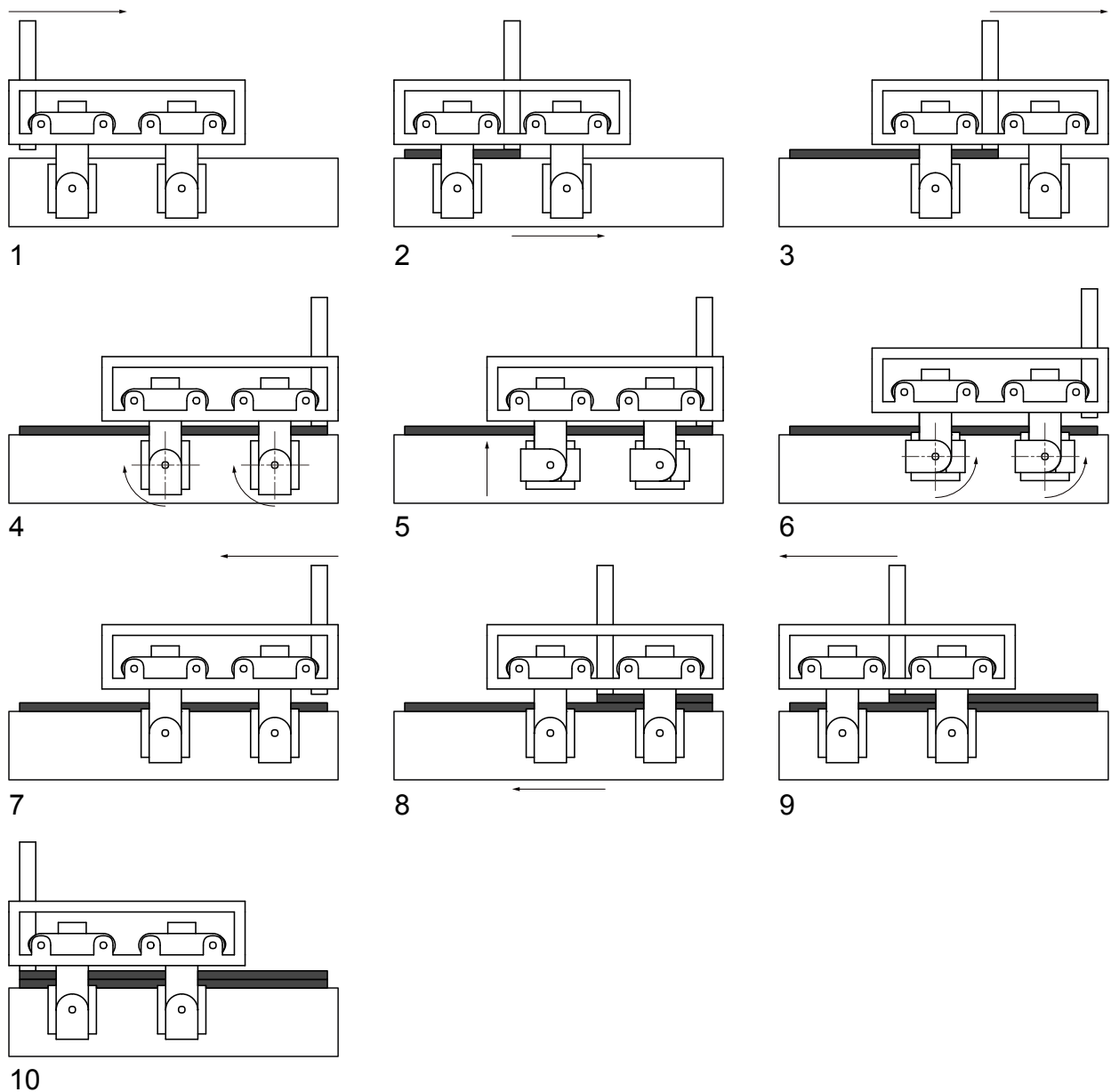

## Print-head shifting

Having the print-head fixed in a fixed position would cause a number of problems, on any curved path the print-head center would deviate from the center of the path as shown in Figure 7, where the center position of the nozzle is marked with a dashed line. In this case each preceding path would deviate accumulatively from the desired shape thus printing a different structure than programmed.

This can be solved by introducing linear side to side motion to the print-head, it can be used to correct the deviation and control the path of the print head illustrated in the hatched section of Figure 7. To position the print-head correctly a sensor calculates the deviation distance and adjusts accordingly. If the position of the device and path curvature are known - deviation can be precalculated geometrically without the need for additional sensors.

If the print-head was confined to only following the previous layer, printed structures could only be shaped as extruded versions of their footprints. This would exclude forms such as vaults, and cantilevers. Shifting the print-head provides an opportunity to alter the inclination of the wall during the printing process as shown in Figure 8. When layers shift relative to their previous layers the inclination of the wall changes. The amount of shift should be precalculated, varying dependent on the position of the device. In the case that the position of the robot is incorrect or an error is detected - curve deviation should be taken into account and either added or subtracted from the shift.

The position of the robot in relation to previous layers can be abstracted in various ways, for example different types of sensors from Local Positioning Systems to rotation counters attached on the wheels leading to many possibilities exhibited in Figure 9.

Fig. 7 Print-head shifts to maintain curvature of the path

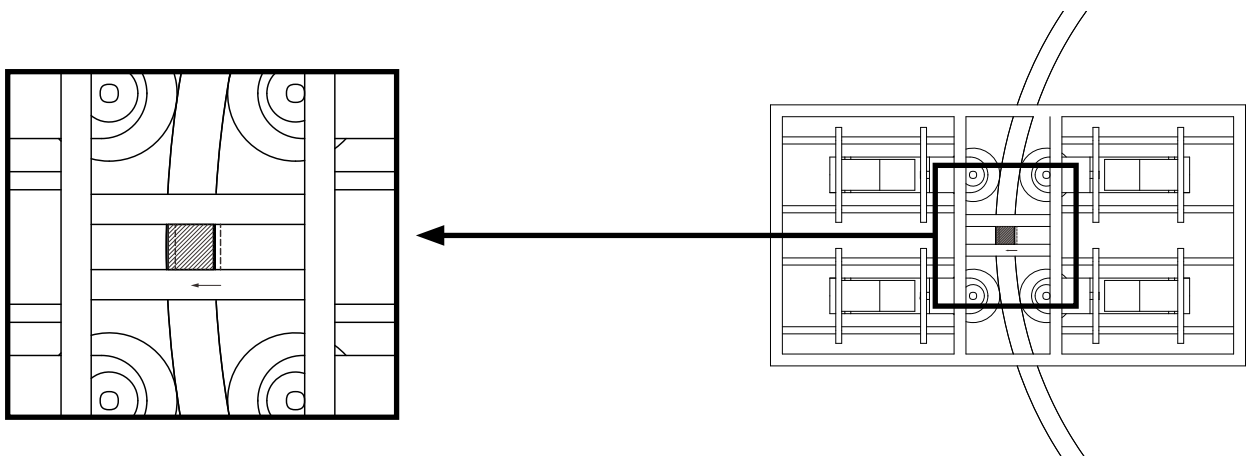

Fig. 8 Print-head shifts to change wall curvature

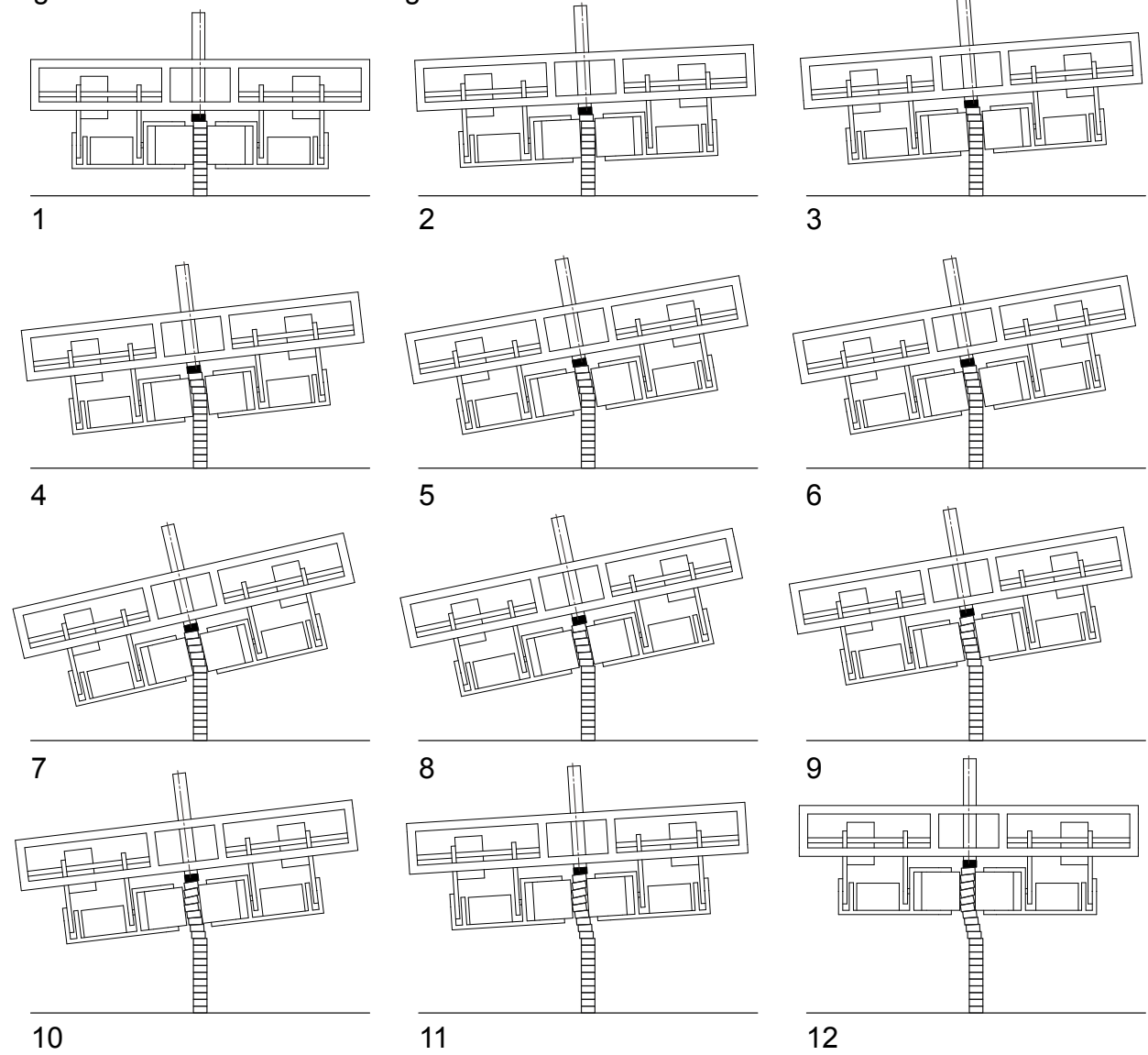

Fig. 9 Printed structures with various wall curvature

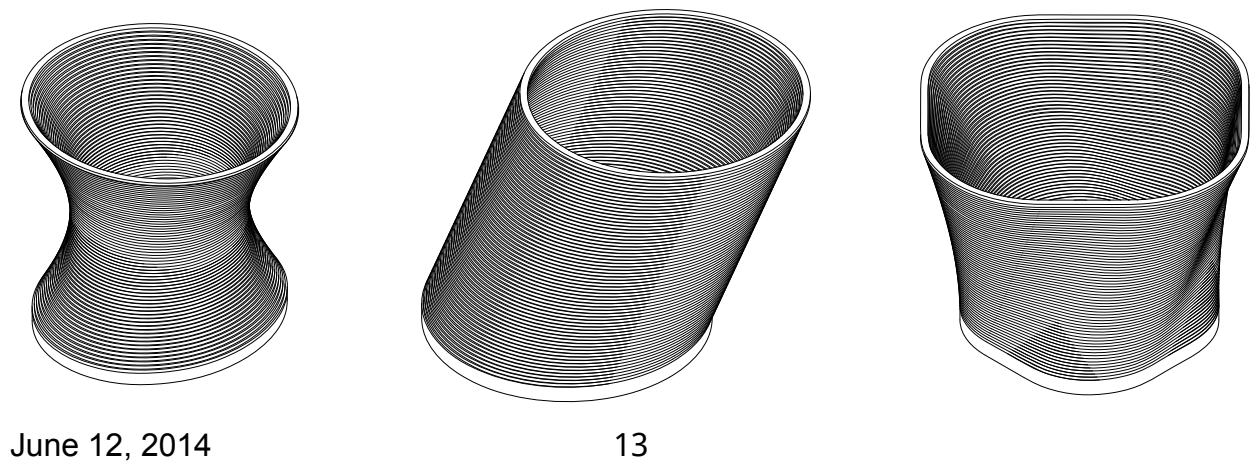